\documentclass[a4paper, times, 10pt,twocolumn]{article}
\usepackage[top=4.9cm,bottom=3.7cm,left=1.5cm,right=1.5cm]{geometry}
\usepackage{ICMLC}
\usepackage{times}
\usepackage{amsmath}
\usepackage{caption}
\usepackage{latexsym}
\usepackage{graphicx}
\usepackage{hyperref}
\usepackage{subcaption}
\usepackage{indentfirst}
\usepackage{amsmath,amssymb,amsfonts}
\usepackage{textcomp}
\usepackage{xcolor}
\usepackage{xspace}
\usepackage{booktabs,multirow}

\newcommand{\myparagraph}[1]{\noindent \textbf{#1}}

\usepackage[margin=8pt,font=footnotesize,labelfont=bf,labelsep=period
]{caption}

\pdfpagewidth=\paperwidth
\pdfpageheight=\paperheight
\begin{document}

\title{SAMPLES ON THIN ICE: RE-EVALUATING ADVERSARIAL PRUNING OF NEURAL NETWORKS}


\author{\bf{\normalsize{GIORGIO PIRAS$^{1,2}$, MAURA PINTOR${^1}$, AMBRA DEMONTIS${^1}$, BATTISTA BIGGIO${^1}$}}\\ 
\\
\normalsize{$^1$Department of Electrical and Electronic Engineering, University of Cagliari, Italy}\\
\normalsize{$^2$Department of Computer Engineering, Sapienza University of Rome, Italy} \\
\normalsize{E-MAIL: giorgio.piras@unica.it, maura.pintor@unica.it, ambra.demontis@unica.it, battista.biggio@unica.it }\\
\\}

\maketitle \thispagestyle{empty}

\begin{abstract}
   Neural network pruning has shown to be an effective technique for reducing the network size, trading desirable properties like generalization and robustness to adversarial attacks for higher sparsity. Recent work has claimed that adversarial pruning methods can produce sparse networks while also preserving robustness to adversarial examples. 
   In this work, we first re-evaluate three state-of-the-art adversarial pruning methods, showing that their robustness was indeed overestimated. We then compare pruned and dense versions of the same models, discovering that samples on thin ice, i.e., closer to the unpruned model's decision boundary, are typically misclassified after pruning. We conclude by discussing how this intuition may lead to designing more effective adversarial pruning methods in future work.
\end{abstract}
\begin{keywords}
   {Machine Learning; Adversarial Examples; Adversarial Robustness; Neural Network Pruning.}
\end{keywords}

\section{Introduction}

Neural network pruning originates in the early 90s \cite{DBLP:conf/nips/CunDS89} but gained notable popularity only after the rise of deep learning, which featured deeper and wider networks over time. Following such dimensional escalation, pruning turned out to be a valuable tool to support the deployment of deep networks in resource-constrained scenarios, where only smaller networks are suited to fit the needs. When operating in such scenarios, pruning a network (i.e., reducing the size by zeroing the parameters) can produce, if properly enhanced, highly sparse yet reasonably good systems, both in terms of generalization \cite{DBLP:conf/mlsys/BlalockOFG20} and robustness to adversarial attacks \cite{DBLP:conf/nips/GuiWYYW019, DBLP:conf/nips/Sehwag0MJ20, DBLP:conf/iccv/YeLX0CLZZMW19}.  
Learning algorithms, in fact, are susceptible to adversarial attacks \cite{biggio13-ecml,goodfellow15-iclr}, where carefully crafted input samples are conceived to fool the classification made by the algorithm. Following the rise of deep learning, both pruning and adversarial machine learning evolved and ultimately crossed paths \cite{sehwag_towards19}.

Neural network pruning is mainly regarded as zeroing parameters (weights and, less frequently, biases) or directly removing network structures of an already pre-trained network, which then needs to be fine-tuned accordingly \cite{DBLP:conf/mlsys/BlalockOFG20}. Depending on the kind of pruning structure, criterion, rate, and pipeline, many approaches combining sparsity and robustness can be found in the literature, all ultimately trying to obtain a sparse and robust network. We call these methods as Adversarial Pruning (AP) methods. Among the most prominent AP works, the contributions of~\cite{DBLP:conf/nips/GuiWYYW019, DBLP:conf/nips/Sehwag0MJ20, DBLP:conf/iccv/YeLX0CLZZMW19} have reached notable results in terms of robustness at considerably high sparsity and with great technique diversity. In~\cite{DBLP:conf/nips/GuiWYYW019}, the authors proposed an \textit{Adversarially Trained Model Compression} technique (\textit{ATMC}) consisting of a unified framework integrating the robustness objective (obtained with adversarial training) to model compression (in the form of pruning factorization and quantization).  \textit{HYDRA}~\cite{DBLP:conf/nips/Sehwag0MJ20} was proposed as an optimization mechanism where the pruning technique is aware of the robust training goal, thus pruning connections based on optimizing the robustness objective. Finally, in \cite{DBLP:conf/iccv/YeLX0CLZZMW19}, the authors leveraged the \textit{Alternating Direction Method of Multipliers (ADMM)} to build a concurrent adversarial training and pruning framework. All proposed approaches reported notable accuracy values on clean and adversarial samples. 
However, there has been a growing concern about the evaluation methods employed to assess robustness, and the research community has thus proposed new frameworks to avoid overestimating robustness. Yet, the SoA AP networks have not been able to address robustness in such a standardized way.

In this work, we use the ensemble of attacks proposed in AutoAttack~\cite{croce20-autoattack} to re-evaluate the AP robustness and offer a much more thorough evaluation. Our contribution in Sect.~\ref{sect:re-eval} shows that the SoA AP papers are overestimating the robustness of their networks. Then, focusing on how the pruned models differ from their dense counterparts, in Sect.~\ref{sect:sample_wise} we show that the samples misclassified (or occasionally corrected) by the pruned model lie in the proximity of the decision boundary.

\section{Re-evaluating and Inspecting Adversarial Pruning Methods}\label{sect:contributions}
In this section, we discuss the lacks that can be encountered in SoA AP methods. Then, we discuss how, with this work, we propose to make up for them and gain insights into why pruned models behave differently. 

\myparagraph{Adversarial Pruning.} 
When the research community discovered that learning algorithms are susceptible to adversarial examples, a massive research wave pursued the study of such phenomenon and never stopped progressing ever since, both in terms of attacks and defenses \cite{biggio18}. While many attacks have gained notable popularity \cite{goodfellow15-iclr, madry18-iclr}, on the defense side, Adversarial Training (AT, \cite{madry18-iclr}) represents so far the quintessential attempt of a defense against adversarial attacks. In Adversarial training, the expressed goal is to train the model to be robust to adversarial examples through a min-max optimization, where the loss is minimized (outer problem) for its worst case (inner problem). It follows that the AT formulation can be integrated into most of the pruning pipeline steps, either when pre-training the model, pruning or fine-tuning, leading over the years to the design of a great and diverse fashion of AP methods, hence models that even if pruned are designed to keep up adversarial robustness.

\myparagraph{Robustness Re-evaluation.} Many attacks proposed in the last years may lead to overestimating robustness when used to perform a security evaluation. Therefore, the authors of \cite{croce20-autoattack} proposed a framework called AutoAttack, which ensemble various parameter-free attacks. This framework has become the minimal and adequate robustness test to foster a more reliable analysis. 
However, we observe that prominent AP methods such as \cite{DBLP:conf/nips/GuiWYYW019, DBLP:conf/nips/Sehwag0MJ20, DBLP:conf/iccv/YeLX0CLZZMW19}, have not yet been able to properly keep up with a standardized and extensive robustness evaluation such as the one proposed with AutoAttack, thus limiting the reliability of their robustness evaluation. For this reason, we consider here a comparable and accountable robustness re-evaluation of SoA AP methods made through AutoAttack. The final goal of this re-evaluation is to fairly compare the robustness of the models obtained by the SoA AP methods. The results highlight a general tendency to mask a lower robustness value and show how robustness drops as the network's sparsity increases. 

\begin{figure}[t]
     \centering
     \begin{subfigure}[b]{0.24\textwidth}
         \centering
         \includegraphics[width=\textwidth]{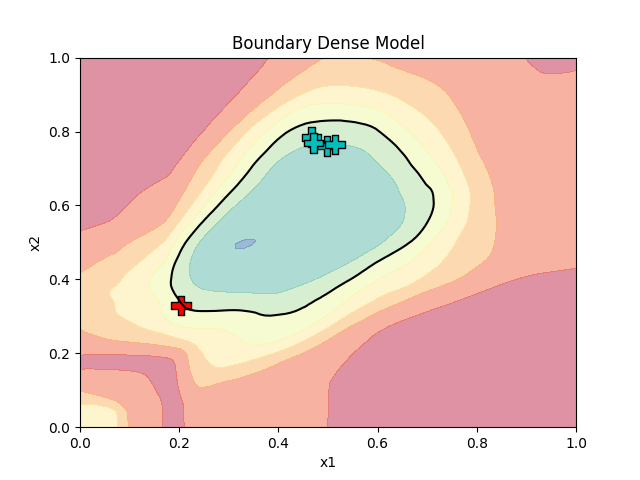}
         \caption{}
         \label{fig:boundary0}
    \end{subfigure}
    \hfill
    \begin{subfigure}[b]{0.24\textwidth}
         \centering
         \includegraphics[width=\textwidth]{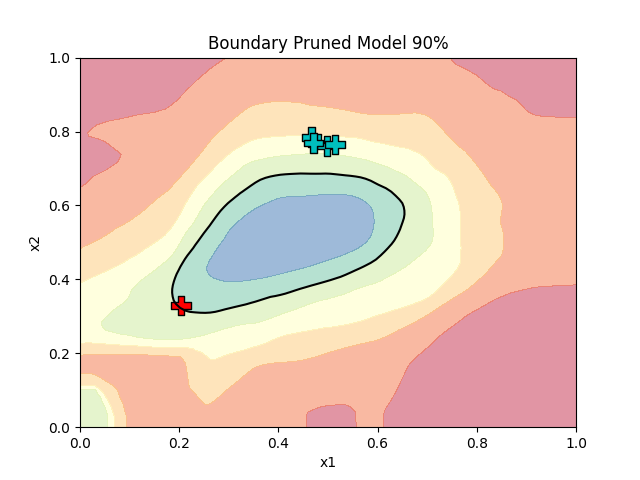}
         \caption{}
         \label{fig:boundary90}
     \end{subfigure}
        \caption{Decision boundary of CIFAR-10 HYDRA WideResNet-28-4 for dense (a) and pruned model at 90\% sparsity (b). The plot contains two sets of samples of the same class whose predictions change as the model gets pruned. In cyan, samples correctly classified by the dense model and later missed after pruning ($\mathcal S_{1,0}$). In red, a sample incorrectly classified by the dense model and later corrected by the pruned ($\mathcal S_{0,1}$).}
        \label{fig:boundary090}
\end{figure}

\myparagraph{Samples on Thin Ice.} We analyzed at the sample level the changes affecting the pruned model with respect to its dense counterpart. Hence, we additionally focused on single samples to understand what changes in the network after pruning make some samples misclassified or corrected by the pruned model, albeit conversely predicted by the dense version. We will refer to these populations as $\mathcal S_{dp}$, where $d,p \in{0,1}$ represent the prediction of the dense (d) and pruned (p) model, which equal 1 if correct and 0 otherwise. As such, the four populations are distributed as follows: 
\begin{itemize}
    \item $\mathcal S_{1,0}$ samples, classified correctly by the dense model and wrongly by the pruned model.
    \item $\mathcal S_{1,1}$ samples, classified correctly by both models. 
    \item $\mathcal S_{0,0}$ samples, classified wrongly by both models. 
    \item $\mathcal S_{0,1}$ samples, classified wrongly by the dense model and corrected by the pruned.
\end{itemize} 
Interestingly, as Figure~\ref{fig:boundary090} shows, we observe that samples lying in the proximity of the decision boundary (samples on thin ice) are more likely to be missed (or occasionally corrected) by the pruned model with respect to the dense, hence more likely to be $\mathcal S_{1,0}$ (or occasionally $\mathcal S_{0,1}$) samples. 

\section{Experimental Analysis}

This section presents the experimental setting, the evaluations and the results for the contributions introduced in Sect.~\ref{sect:contributions}. In Sect~\ref{sect:re-eval}, we illustrate our re-evaluation details, including the AutoAttack setup, and eventually, we show the re-evaluation results of Table~\ref{tab:eval}. Finally, in Sect~\ref{sect:sample_wise}, we focus on the sample-wise differences that we observed in AP models and demonstrate the strong correlation of misclassified samples with their distance to the boundary, as Table~\ref{tab:stats} shows.

\subsection{Robustness Re-evaluation}\label{sect:re-eval}
Through the ensemble of attacks proposed with AutoAttack (AA)~\cite{croce20-autoattack} and the standardized robustness benchmark RobustBench~\cite{croce20-robustbench}, the research community has provided tools and motivation to produce sounder and more proper robustness evaluations. Therefore, it is fundamental to extend such concepts to the models, such as AP ones, that lack this perspective. Aside from HYDRA~\cite{DBLP:conf/nips/Sehwag0MJ20}, which instantiated an insufficient AutoAttack evaluation and uploaded a WideResNet-28-10 model to RobustBench, none of the other methods has pursued a suitable evaluation of their robustness. In both ATMC \cite{DBLP:conf/nips/GuiWYYW019} and ADMM \cite{DBLP:conf/iccv/YeLX0CLZZMW19}, the methods have been tested on PGD attack at best~\cite{madry18-iclr}.

\subsubsection{Experimental Setup}
We narrowed down the analysis to what we consider to be three of the most prominent AP works; ATMC ~\cite{DBLP:conf/nips/GuiWYYW019}, HYDRA ~\cite{DBLP:conf/nips/Sehwag0MJ20} and ADMM ~\cite{DBLP:conf/iccv/YeLX0CLZZMW19}. All the experiments were made using two networks for each method. For each network, we pruned the models to two final sparsity rates, 90\% and 95\%, leading the overall combination to a total of twelve AP models.

\myparagraph{Dataset.}
We used the CIFAR10 dataset to test each of these combinations. The networks were trained and fine-tuned on 50k samples. The evaluation stage, for both re-evaluation and sample-wise analysis (apart from the Mann-Whitney U test), involved the same 1k samples for each model. The size of this portion of the test set is deemed enough to ultimately bring similar results in terms of both accuracy and robustness. Additionally, it helps us provide a more compact visualization of the analyzed sample-wise phenomenon.

\myparagraph{Models.} 
For every method, we have chosen two specific networks among VGGs, ResNets, and WideResNets. The availability of each library has trivially oriented the choice of the models. Specifically, for the ATMC method, the two selected networks are ResNet-34 and ResNet-18; for ADMM, a ResNet-18 and a VGG-16; finally, for the HYDRA method, a WideResNet-28-4 and a VGG-16.

\myparagraph{Attack.}
The employed version of the AutoAttack framework~\cite{croce20-autoattack} is \textit{plus}, which includes untargeted APGD-CE (5 restarts), untargeted APGD-DLR (5 restarts), untargeted FAB (5 restarts), Square Attack (5000 queries), targeted APGD-DLR (9 target classes), targeted FAB (9 target classes). APGD, contrary to the classic formulation, is based on a variable step size that adapts to the budget and the progress of the optimization. The other attacks are a selection of minimum norm and black-box attacks to guarantee a wider attack spectrum \cite{croce20-fab-icml, DBLP:conf/eccv/AndriushchenkoC20}.  

\subsubsection{Re-Evaluation Results}
The following subsection shows the results collected from re-evaluating the models. In Table~\ref{tab:eval}, we show accuracy and robustness for every method/network/sparsity combination under the AutoAttack evaluation. 
\begin{table*}[h]
\footnotesize\centering
\caption{Robustness and Accuracy (\%) after re-evaluating existing methods for pruning and robustness. In column \textit{\textbf{Rep.Rob.}} we indicate the reported clean accuracy indicated either in the paper or after training the model with their own library if not reported in the paper or reported for a different sparsity ratio. In column \textit{\textbf{A.A. Rob.}}  we report the robust accuracy evaluated with AutoAttack \cite{croce20-autoattack} on a 1000-sample test set.}
\begin{tabular}{|l|l|l|l|l|l|l|l|}
\hline
\textbf{Method} & \textbf{Sparsity \%} & \textbf{Network} & \textbf{Rep. Acc. \%} & \textbf{A.A. Acc. \%} & \textbf{Rep. Rob. \%} & \textbf{A.A. Rob. \%} & \textbf{Drop \%} \\ \hline
ATMC            & 90          & ResNet-34        & 84.83            & 85.60            & 43.78            & 38.20            & 5.58         \\ \hline
ATMC            & 95          & ResNet-34        & 84.71            & 84.90            & 43.24            & 36.30            & 6.94         \\ \hline
ATMC            & 90          & ResNet-18        & 84.91            & 84.90            & 41.83            & 36.90            & 2.23         \\ \hline
ATMC            & 95          & ResNet-18        & 84.47            & 84.60            & 42.75            & 35.70            & 7.05         \\ \hline
ADMM            & 90          & ResNet-18        & 81.43            & 81.30            & 43.37            & 44.07            & -0.70        \\ \hline
ADMM            & 95          & ResNet-18        & 80.19            & 80.30            & 44.17            & 43.00            & 1.17         \\ \hline
ADMM            & 90          & VGG-16           & 77.19            & 76.10            & 43.56            & 40.50            & 3.60         \\ \hline
ADMM            & 95          & VGG-16           & 73.14            & 73.20            & 43.96            & 38.20            & 5.76         \\ \hline
HYDRA           & 90          & WRN28-4          & 83.70            & 84.60            & 55.20            & 52.80            & 2.40         \\ \hline
HYDRA           & 95          & WRN28-4          & 82.70            & 84.10            & 54.20            & 50.90            & 3.30         \\ \hline
HYDRA           & 90          & VGG-16           & 80.50            & 81.50            & 49.50            & 46.70            & 2.80         \\ \hline
HYDRA           & 95          & VGG-16           & 78.90            & 79.30            & 48.70            & 45.10            & 3.60         \\ \hline
\end{tabular}
\label{tab:eval}
\end{table*}
The reported (Rep.) values of Table~\ref{tab:eval} are the results listed in the paper or, alternatively, given by the library itself after the training has been completed. The re-evaluated robustness, exception made for the ADMM ResNet-18, is always lower. In the case of ATMC, we highlight a minimum robustness drop of 2.23\% for the ResNet-18 pruned at 90\% sparsity, while the remaining 3 networks have a rather severe decrease, even up to 7.05\% for the ResNet-18 pruned at 95\% sparsity. It is worth highlighting that the networks pruned to the highest sparsity always register the biggest drop. In the ADMM case, the reported value tends to be accurate for the ResNet-18 case, while the VGG-16, likely due to poor baseline performances, has a considerable drop in the 95\% case. We found HYDRA to have the least volatile drop throughout the combinations yet not reaching negligible differences, unlike the ADMM ResNet-18 case, which marks the top of the list regarding lower drop. The HYDRA WideResNet-28-4 is instead consistently the top-performing model overall. 

\subsection{Samples on Thin Ice}\label{sect:sample_wise}
In Section~\ref{sect:re-eval}, we analyzed the robustness value of AP networks measured with AutoAttack showing how, in most cases, the methods were overestimating the robustness to adversarial examples and how this overestimation was more severe in some cases. Yet, a still challenging question follows the analysis of pruned models: \textit{How does a network's boundary change after pruning? On what samples will it introduce more errors?} 
In a security-related scenario where these AP networks are supposed to operate, it is crucial to understand how the model has changed its predictions and why. Additionally, gathering insights into the pruned model from the past dense model is more challenging yet intriguing. Driven by these questions, we try to understand what characterizes the samples on which the two versions of the same model disagree. 

\subsubsection{Samples and Distances}
\begin{figure*}[h]
     \centering
     \begin{subfigure}[b]{0.33\textwidth}
         \centering
         \includegraphics[width=\textwidth]{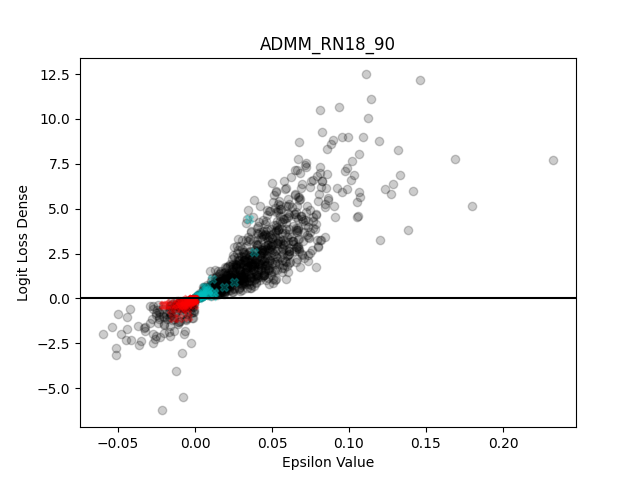}
         \caption{}
         \label{fig:admmrn1890}
    \end{subfigure}
    \hfill
    \begin{subfigure}[b]{0.33\textwidth}
         \centering
         \includegraphics[width=\textwidth]{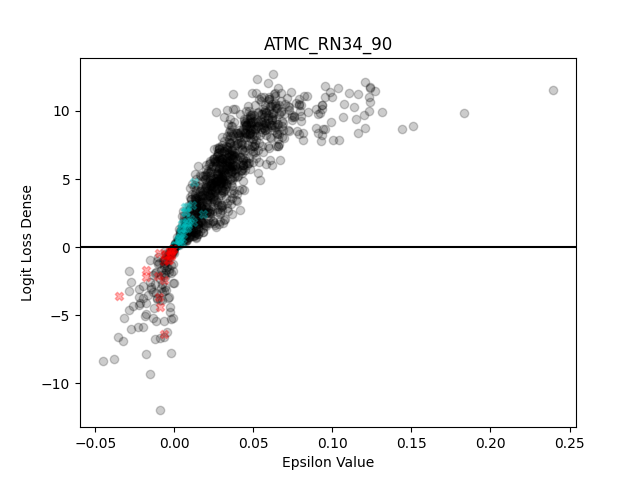}
         \caption{}
         \label{fig:atmcrn3490}
     \end{subfigure}
     \hfill
     \begin{subfigure}[b]{0.33\textwidth}
         \centering
         \includegraphics[width=\textwidth]{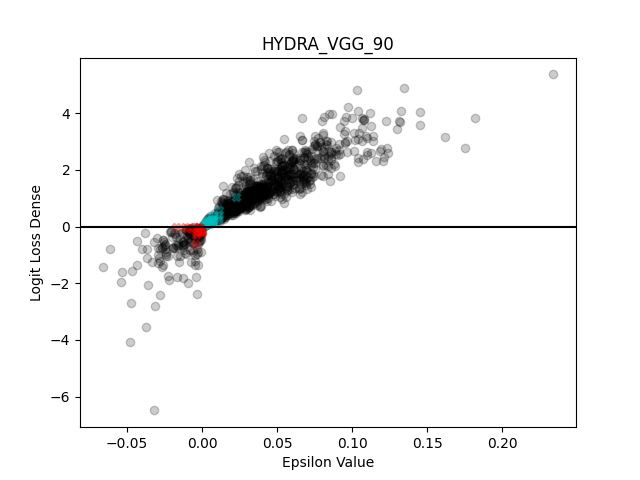}
         \caption{}
         \label{fig:hydravgg90}
    \end{subfigure}
        \caption{Loss of the dense model versus $\epsilon$ for one network of each method at 90\% sparsity. The black dots over the 0 thresholds represent the samples in ($\mathcal S_{1,1}$), while the cyan crosses represent the samples in ($\mathcal S_{1,0}$). The black dots below the 0 thresholds represent the samples in ($\mathcal S_{0,0}$), while the red crosses represent the samples in ($\mathcal S_{0,1}$). Being the distance to the boundary $\epsilon$ equal to 0 for samples misclassified by the dense model (for $\mathcal S_{0,0}$ and $\mathcal S_{0,1}$ samples, the minimum distance to make a sample adversarial measured by FMN is 0 as the class is predicted wrongly), we still measure the distance to the boundary by imposing the target class as predicted and underline it through a negative $\epsilon$.}
        \label{fig:eps_vs_logitloss}
\end{figure*}

Our basic intuition is that pruning a model from mild to high sparsities imports slight variations on the decision boundary of the network, as also reported in past works \cite{DBLP:journals/pr/YeomSLBWMS21}. Following these observations, we applied the Fast-Minimum-Norm (FMN, \cite{pintor2021fast}) attack to the dense model, which enabled us to empirically assess the distance to the decision boundary of each test sample, which we will refer to as epsilon ($\epsilon$). Being the decision boundary affected by small changes after pruning, we conjectured that $\mathcal S_{1,0}$ and $\mathcal S_{0,1}$ samples were merely a portion of samples near the decision boundary and thus more likely to be misclassified or corrected after pruning. Figure~\ref{fig:eps_vs_logitloss} shows for one network pruned at 90\% sparsity of each of the considered methods, the scatter of each $S-$~population along the logit loss of the dense model versus the epsilon value. We recall that the logit loss is negative for misclassified samples and positive for correctly classified ones, hence growing positively if confidence grows \cite{carlini17-sp}. 
 As shown in Figure~\ref{fig:eps_vs_logitloss}, the $\mathcal S_{0,1}$ and $\mathcal S_{1,0}$ samples, represented by red and cyan crosses respectively, lie in the surroundings of the decision boundary, confirming our hypothesis that both positive and negative flips in the pruned-dense comparison concern the distance to the boundary. It is yet crucial to understand how strong the correlation of these groups of samples with epsilon is in reality. Given that the for $\mathcal S_{d,p}$ populations can be subdivided by the dense outcome, we are interested in knowing how the $\epsilon$ value is significant in separating $\mathcal S_{1,0}$ from $\mathcal S_{1,1}$, and finally $\mathcal S_{0,1}$ from $\mathcal S_{0,0}$.

\begin{table*}[h]
\footnotesize
\centering
\caption{The stats for each considered model identified by method, sparsity rate and used network. In columns \textbf{\textit{iv}} to \textbf{\textit{vii}}, the values represent the percentage of samples belonging to each population. Columns \textbf{\textit{viii}} and \textbf{\textit{ix}}, through the AUC values, report the probability that a randomly chosen sample from one distribution has a lower $\epsilon$ value than a randomly chosen sample from the other distribution. The p-value made with the Mann-Whitney U Test is also reported in brackets, indicating how separable the two distributions can be based on $\epsilon$. The tests were run for both ``macro-populations" $\mathcal S_{1,p}$ and $\mathcal S_{0,p}$, where $p \in{0,1}$.}
\begin{tabular}{|l|l|l|l|l|l|l|l|l|l|l|l|}
\hline
\textbf{Method} & \textbf{Sparsity \%} & \textbf{Network} & \textbf{$\mathcal S_{0,0}$\%} & \textbf{$\mathcal S_{0,1}$\%} & \textbf{$\mathcal S_{1,0}$\%} & \textbf{$\mathcal S_{1,1}$\%} & \textbf{AUC$_{\epsilon_{10}<\epsilon_{11}}$(p)} & \textbf{AUC$_{|\epsilon_{01}|<|\epsilon_{00}|}$(p)}\\ \hline
ATMC            & 90         & ResNet-34        & 11.45          & 2.65           & 3.43              & 82.47 & 0.941 (4e-170) &   0.679 (4e-20)      \\ \hline
ATMC            & 95          & ResNet-34        & 11.09          & 3.01           & 4.44              & 81.46 & 0.927 (1e-203) & 0.665 (5e-19)      \\ \hline
ATMC            & 90          & ResNet-18        & 12.31          & 4.54           & 3.63              & 79.52 & 0.908 (8e-154) & 0.659 (3e-24)       \\ \hline
ATMC            & 95          & ResNet-18        & 12.4           & 4.45           & 3.90              & 79.25 & 0.894 (1e-153) & 0.639 (1e-18)     \\ \hline
ADMM            & 90          & ResNet-18        & 15.52          & 5.04           & 3.71              & 75.73 & 0.896 (3e-147) & 0.669 (9e-31)     \\ \hline
ADMM            & 95          & ResNet-18        & 15.98          & 4.58           & 4.53              & 74.91 & 0.894 (5e-176) & 0.662 (1e-26)      \\ \hline
ADMM            & 90          & VGG-16           & 20.88          & 4.29           & 4.24              & 70.59 & 0.912 (7e-180) & 0.691 (3e-36)    \\ \hline
ADMM            & 95          & VGG-16           & 21.17          & 4.0            & 5.98              & 68.85 & 0.914 (2e-248) & 0.676 (2e-29)    \\ \hline
HYDRA           & 90          & WRN28-4          & 13.14          & 1.51           & 3.17              & 82.18 & 0.955 (2e-167) & 0.768 (1e-27)     \\ \hline
HYDRA           & 95          & WRN28-4          & 13.44          & 1.21           & 3.89              & 81.46 & 0.952 (1e-200) & 0.782 (2e-25)    \\ \hline
HYDRA           & 90          & VGG-16           & 15.85          & 1.43           & 3.6               & 79.12 & 0.951 (9e-186) & 0.796 (2e-32)  \\ \hline
HYDRA           & 95          & VGG-16           & 16.2           & 1.08           & 4.91              & 77.81 & 0.944 (2e-240) & 0.713 (5e-14)   \\ \hline
\end{tabular}
\label{tab:stats}
\end{table*}

Table~\ref{tab:stats} provides a meticulous empirical display of the statistics related to the samples and their $\epsilon$ value. First, it shows the distribution of samples across all populations. Then, we carry a statistical analysis of the separability inside $\mathcal S_{1,p}$ (i.e. $\mathcal S_{1,0}$ vs $\mathcal S_{1,1}$) and $\mathcal S_{0,p}$ (i.e. $\mathcal S_{0,1}$ vs $\mathcal S_{0,0}$). The analysis is based on the AUC value, and Mann/Whitney U test p-value, computed on the entire test set of 10k samples to yield a more accurate analysis. In the first case, the AUC value indicates the probability that a randomly chosen sample from $\mathcal S_{1,0}$ has a lower $\epsilon$ value than a randomly selected sample from $\mathcal S_{1,1}$. The p-value, instead, represents the probability that the null hypothesis (indicating that the probability distribution of a randomly drawn observation from $\mathcal S_{1,1}$ is the same as the probability distribution of a randomly drawn observation from $\mathcal S_{1,0}$) is true. The results for both AUC and p-value suggest a very high statistical significance of the distance to the boundary $\epsilon$ with respect to the two $\mathcal S_{1,p}$ populations, thus confidently rejecting the null hypothesis. These findings, overall, suggest that it is possible to significantly estimate from prior information the samples that the pruned model will misclassify with respect to the dense, hence implying that to uplift the performance of the pruned model, we might consider a training procedure weighting samples based on their distance to the boundary.
The results of the remaining two $\mathcal S_{0,p}$ populations still suggest a high statistical significance for both AUC and p-value. 

\section{Related Work}\label{sect:related}

Starting from \cite{sehwag_towards19}, the authors demonstrated that a small and robust model is less robust than a bigger model compressed to the same size and analyzed the relationship between robustness and sparsity. Following these works, several methods focused on AP. Yet, we have limited the choice to three of the most valuable works according to the results achieved in terms of robustness, the claims, the evaluation method and whether the code was published \cite{DBLP:conf/nips/GuiWYYW019, DBLP:conf/nips/Sehwag0MJ20, DBLP:conf/iccv/YeLX0CLZZMW19}.

Regarding the sample-wise analysis of Section~\ref{sect:sample_wise}, other works have tried to characterize the difference between pruned and dense models as in \cite{DBLP:conf/mlsys/LiebenweinB00R21}, which focused on the similarity between the models, their robustness to noise and to out-of-distribution samples. Unlike this approach, we base our analysis on adversarial attacks and, additionally, we focus on samples to characterize what changes between the models and why. In spirit, our sample-wise analysis is similar to what has been proposed in \cite{hooker_whatdo_nns_forget}, where the authors coined the term Pruning Identified Exemplars (PIEs) to identify the subset of samples more impacted by compression, which the authors concluded were mostly rare and atypical examples. Following \cite{hooker_whatdo_nns_forget}, a series of works tried to analyze whether the pruned models were impacted adversely while representing the least prevalent groups of samples (e.g. classes) in imbalanced datasets, hence from a fairness perspective such as in \cite{DBLP:journals/tran/abs-2205-13574}. Upon finding positive evidence of such an effect for imbalanced datasets, the authors correlated the more impacted groups (i.e. smaller groups in size) with bigger gradient norms and group Hessians, with the latter being bounded by the groups' distance to the boundary. On the contrary, we analyze why single samples in a balanced setting are classified differently by pruned and robust neural networks. We conclude that samples whose predictions disagree with dense and pruned models lie nearby the boundary. Additionally, we empirically demonstrate how the distance characterizes the samples' populations statistically.

\section{Conclusions and Future Work}
In this work, we have shown that the robustness of the current SoA AP models is often overestimated. Through a unified analysis of such methods and in a comparable setting, we have re-evaluated the models presenting a more thoughtful security evaluation.
Additionally, we have developed a sample-wise analysis aiming to characterize the difference between dense and pruned robust models, featuring a careful statistical analysis of the samples' distance to the decision boundary to establish and ultimately confirm the validity of such correlation. Additionally, for future work, our results on the statistical behaviour of the distance to the boundary suggest exploring a training procedure to uplift AP models' performances based on the distance of a sample from the boundary.

\section*{Acknowledgements}
This work has been carried out while G. Piras was enrolled in the Italian National Doctorate on Artificial Intelligence run by the Sapienza University of Rome in collaboration with the University of Cagliari, and it has been partly supported by the PRIN 2017 project RexLearn (grant no. 2017TWNMH2), funded by the Italian Ministry of Education, University and Research; by BMK, BMDW, and the Province of Upper Austria in the frame of the COMET Programme managed by FFG in the COMET Module S3AI; and by project SERICS (PE00000014) under the MUR National Recovery and Resilience Plan funded by the European Union – NextGenerationEU.

{\small
\bibliographystyle{abbrv} 

}

\end{document}